\pdfoutput=1
\PassOptionsToPackage{table, dvipsnames}{xcolor}

\documentclass[11pt]{article}

\usepackage{xcolor}
\usepackage[final]{acl}

\usepackage{times}
\usepackage{latexsym}
\usepackage{amsfonts}

\usepackage[T1]{fontenc}

\usepackage[utf8]{inputenc}

\usepackage{microtype}

\usepackage{inconsolata}

\usepackage{graphicx}
\usepackage{amsmath}
\usepackage{enumitem}
\usepackage{booktabs}
\usepackage{adjustbox}
\usepackage{multirow}

\newcommand{\ourmethod}{{\textsc{OPPU}}}
\usepackage{adjustbox}
\usepackage{makecell}
\usepackage{enumitem}

\title{Democratizing Large Language Models via\\ Personalized Parameter-Efficient Fine-tuning}

\author{Zhaoxuan Tan$^{\clubsuit}$, Qingkai Zeng$^{\clubsuit}$, Yijun Tian$^{\clubsuit}$, Zheyuan Liu$^{\clubsuit}$, Bing Yin$^{\diamondsuit}$, Meng Jiang$^{\clubsuit}$\\
        $^{\clubsuit}$University of Notre Dame, $^{\diamondsuit}$Amazon.com Inc.\\
        \texttt{\{ztan3, mjiang2\}@nd.edu}}

\begin{document}
\maketitle

\begin{abstract}
Personalization in large language models (LLMs) is increasingly important, aiming to align the LLMs' interactions, content, and recommendations with individual user preferences.
Recent advances have highlighted effective prompt design by enriching user queries with non-parametric knowledge through behavior history retrieval and textual profiles. However, these methods faced limitations due to a lack of model ownership, resulting in constrained customization and privacy issues, and often failed to capture complex, dynamic user behavior patterns.
To address these shortcomings, we introduce \textbf{O}ne \textbf{P}EFT \textbf{P}er \textbf{U}ser (\ourmethod{})\footnote{The code is available at \url{https://github.com/TamSiuhin/OPPU}}, employing personalized parameter-efficient fine-tuning (PEFT) modules to store user-specific behavior patterns and preferences. By plugging in personal PEFT parameters, users can own and use their LLMs individually. \ourmethod{} integrates parametric user knowledge in the personal PEFT parameters with non-parametric knowledge from retrieval and profiles, adapting LLMs to user behavior shifts. Experimental results demonstrate that \ourmethod{} significantly outperforms existing prompt-based methods across seven diverse tasks in the LaMP benchmark. Further studies reveal \ourmethod{}'s enhanced capabilities in handling user behavior shifts, modeling users at different activity levels, maintaining robustness across various user history formats, and displaying versatility with different PEFT methods.

\end{abstract}

\section{Introduction}
Personalization refers to mining users' behavior history, and therefore tailoring and customizing a system's interactions, content, or recommendations to meet specific needs, preferences, and characteristics of individual users \cite{tan2023user, chen2023survey}. By adapting to each user's preferences, personalization systems enhance user experience, increasingly getting vital in areas like content recommendation \cite{qian2013personalized, wu2023personalized, baek2023knowledge}, user simulation \cite{dejescu2023approaches}, personalized chatbots \cite{srivastava2020personalized, ma2021one}, user profiling \cite{gu2020hierarchical, gao2023chat}, healthcare \cite{goldenberg2021personalization}, and education \cite{pratama2023revolutionizing}.

\begin{figure}[t]
    \centering
    \includegraphics[width=1\linewidth]{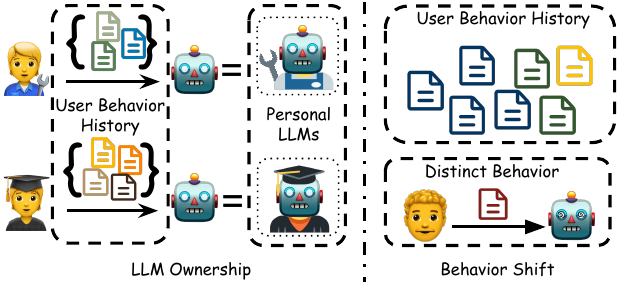}
    \caption{\textit{LLM ownership} and \textit{behavior shift} are two challenges that developing personalized LLMs has to face. Ownership emphasizes that the model needs to be owned by individual user to enhance customization and privacy. Behavior shift adaption refers to the LLMs' ability to effectively generalize and adapt to emerging new patterns in user behaviors.}
    \label{fig:teaser}
\end{figure}

Large language models (LLMs) display emergent abilities not seen in smaller models \cite{wei2022emergent, lu2023emergent}, as they have billions of parameters and are trained on vast corpora.
However, existing LLMs predominantly follow the ``one-size-fits-all'' paradigm.  They are generally trained on extensive, domain-agnostic datasets, which limits their effectiveness in meeting the specific needs and preferences of individual users \cite{chen2023large}. Therefore, the challenge of integrating the strong generative capabilities of LLMs with the tailored requirements of individual users has emerged as a significant area of research \cite{li2023teach}.

Existing works on personalizing LLMs have predominantly concentrated on developing prompt templates, which fall into three categories: vanilla, retrieval-augmented, and profile-augmented personalized prompts. The vanilla personalized prompt approach leverages the in-context learning capability of LLMs, utilizing the user's entire or randomly sampled history as contextual examples \cite{dai2023uncovering, zhiyuli2023bookgpt}. 
Considering the growing length of user behavior history and the limited LLM context length, some studies applied retrieval methods to select the most relevant part of user behavior history to enhance LLM personalization \cite{mysore2023pearl}. Besides the retrieval, some techniques explicitly generate user preferences and profiles in natural language to augment LLMs' input \cite{richardson2023integrating}.


Despite much research progress has been made in LLM personalization, existing methods face ownership and behavior shift challenges (Fig. \ref{fig:teaser}):
\begin{itemize}[leftmargin=*]
    \item \textbf{Ownership}: Existing methods are processed centralized, where user history is encoded in a personalized prompt and processed by centralized LLMs. This paradigm limits the model's customization and ability to provide deep, personalized experiences tailored to individual users. Moreover, when using a centralized model, users often have to share personal data with the service provider, which raises concerns about how user data are stored, used, and protected.
    \item \textbf{Behavior Pattern Generalization}: As is revealed by \citet{shi2023large}, LLMs can be easily distracted by irrelevant context information that retrieval can hardly avoid.
    In LLM personalization, where the retrieval corpus is confined to a specific user's behaviors, retrieval augmentation might underperform, especially when the user's past behaviors do not closely mirror the patterns needed for the query at hand.
\end{itemize}

In light of these challenges, we propose \textbf{O}ne \textbf{P}EFT \textbf{P}er \textbf{U}ser (\ourmethod{}), equipping each user with a personalized, parameter-efficient fine-tuning (PEFT) module. Characterized by PEFT's plug-and-play functionality and the minimal weight of updated parameters (typically less than 1\% of the base LLM), \ourmethod{} facilitates LLM ownership and enhances generalization in scenarios of user behavior shifts. 
By fine-tuning the PEFT module with the user's personal behavior history, the personalized PEFT parameters encapsulate behavior patterns and preferences. This process, when integrated into base LLMs, allows users to obtain their private LLMs, ensuring LLM ownership and enhancing model customization. 
Furthermore, as is revealed by \citet{gupta2024rag}, fine-tuning LLMs is more effective than retrieval augmentation when the retrieved instances are not highly relevant to the query.
The fine-tuned personal LLMs in \ourmethod{} are adept at capturing complex behavior patterns and thus capable of understanding new behaviors with less reliance on highly relevant history data.
Experimental results show that \ourmethod{} outperforms all baselines on seven public tasks in the Language Model Personalization (LaMP) benchmark \cite{salemi2023lamp}. Additional studies emphasize the importance of integrating non-parametric user knowledge from retrieved history with parametric knowledge from personal PEFT parameters. In scenarios of user behavior shifts, where history is less relevant, \ourmethod{} significantly outperforms retrieval-based methods. Moreover, \ourmethod{} is resilient to varying user history formats and demonstrates versatility across different PEFT methods, among other advantages.

To summarize, the contribution of \ourmethod{} lies in its pioneering approach to PEFT-based LLM personalization. Each user (or user cohort) benefits from a personal PEFT module, which not only ensures LLM ownership but also significantly improves the model's ability to adapt to shifts in user behavior. The superiority of \ourmethod{} is evidenced by state-of-the-art performance across seven tasks in the LaMP benchmark. By introducing this innovative parametric-based personalization technique, \ourmethod{} opens up new opportunities in democratizing personalized LLMs. 


\begin{figure*}[t]
    \centering
    \includegraphics[width=0.8\linewidth]{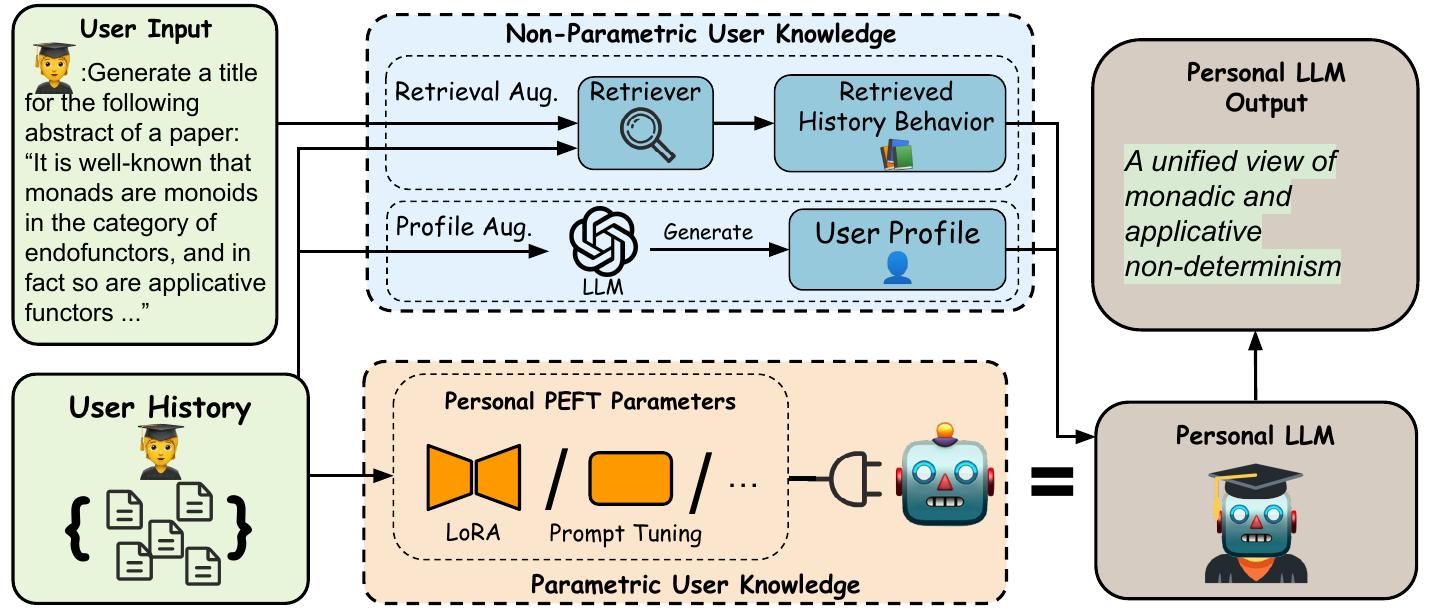}
    \caption{Overview of our proposed \ourmethod{}, where each user is equipped with a personal PEFT module and plug-in base LLMs to get their individual LLM. Beyond parametric personalization via PEFT, \ourmethod{} is also compatible with the non-parametric user knowledge via retrieval and profile augmentation.}
    \label{fig:overview}
\end{figure*}

\section{Preliminaries}

\subsection{Research Problem Formulation}
For personalizing LLMs at time $t$, the output $r_u$ for user $u$ is conditioned on both input $q_u$ and the user’s behavior history $\mathcal{H}_u$. Specifically, $\mathcal{H}_u  = \{h_u\}$, includes all user behaviors $h_{u}$ before time $t$. User behavior $h_u$ may consist of $(x_{u}, y_{u})$ pairs, aligning with the task-specific query-answer format $(q_u, r_u)$, or plain text sequences $x_u$ providing context for behavior patterns. We aim to obtain personalized parameters $\Theta_u$ for each user $u$.


\subsection{Base LLMs Task Adaption}

Given that off-the-shelf LLMs are not inherently equipped for personalization tasks, we follow the methods of LaMP \cite{salemi2023lamp} by fine-tuning LLMs for fair comparison. This section outlines the development of base LLMs with a set of held-out users to enhance their general capabilities for personalization tasks without involving target user preferences. Specifically, we provide three alternatives: base LLM $\Theta^\mathrm{(B)}$ that only involves task related data, retrieval-augmented base LLM $\Theta^\mathrm{(R)}$ that augment input with top-$k$ relevant user history, and profile-augmented base LLM $\Theta^\mathrm{(P)}$ that involves textual user profiles as input.
Note that introducing RAG and PAG means users would expose their historical data or profiles to a centralized LLM, potentially affect the model ownership. For users prioritizing privacy and ownership, \ourmethod{} without retrieval avoids revealing user data to service providers. Conversely, those seeking optimal performance and consent to reveal data to centralized LLMs should opt for RAG or PAG. The fine-tuning objectives of three base models are:
\[
\left\{
\begin{aligned}
    \mathcal{L}_\mathrm{B}=&\mathrm{CE}[\Theta^\mathrm{(B)}(\phi_{t}(q_u)), r_u]\\
    \mathcal{L}_\mathrm{R}=&\mathrm{CE}[\Theta^\mathrm{(R)}(\phi_{r}(q_u, \mathcal{D}_u)), r_u],\\
    \mathcal{L}_\mathrm{P}=&\mathrm{CE}[\Theta^\mathrm{(P)}(\phi_{p}(q_u, \mathcal{D}_u, s_u)), r_u],
\end{aligned}
\right.
\]
where $\mathrm{CE}$ denotes the cross entropy loss function, $\phi_t$, $\phi_r$, and $\phi_p$ denote prompt construction function for base, retrieval-augmented, and profile-augmented LLM. The retrieved user history $\mathcal{D}_u=\mathcal{R}(q_u, \mathcal{H}_u, k)$ denotes the top-$k$ user history from retriever $\mathcal{R}$. $s_u=\mathrm{LLM}(\mathcal{H}_u)$ is a textual user profile generated by an instruction-tuned LLM, \emph{e.g.}, \texttt{Vicuna} \cite{chiang2023vicuna}, based on user history.

To make this process more computationally efficient, we adopt the low-rank adaptation (LoRA) \cite{hu2021lora} for base LLM task adaption that only updates about 0.5\% external parameters compared to the total LLM parameter size. After training, LoRA parameters are merged into the base model, equipping LLMs with task capabilities.

\section{One PEFT Per User (\ourmethod{})}

Once the base model for task adaption is obtained, users can only access the base model parameters and their personal behavior history data, controlling privacy risks. 
This section introduces personalized LLMs for target users through parametric PEFT and integrates non-parametric knowledge such as retrieval and profile augmentation.
For each user, we plug a personal trainable PEFT module (LoRA by default) $\Delta\Theta_u^\mathrm{(B)}$, $\Delta\Theta_u^\mathrm{(R)}$, $\Delta\Theta_u^\mathrm{(P)}$ to corresponding base LLM under three settings to obtain personalized LLM $\Theta_u^\mathrm{(B)}$, $\Theta_u^\mathrm{(R)}$, and $\Theta_u^\mathrm{(P)}$, while base LLM parameters $\Theta^\mathrm{(B)}$, $\Theta^\mathrm{(R)}$, $\Theta^\mathrm{(P)}$ are frozen.
\[
\left\{
\begin{aligned}
    \Theta_u^\mathrm{(B)}=&\Theta^\mathrm{(B)}\oplus\Delta\Theta_u^\mathrm{(B)}, \\
    \Theta_u^\mathrm{(R)}=&\Theta^\mathrm{(R)}\oplus\Delta\Theta_u^{(R)}, \\
    \Theta_u^\mathrm{(P)}=&\Theta^\mathrm{(P)}\oplus\Delta\Theta_u^\mathrm{(P)}.
\end{aligned}
\right.
\]
We then use the user data $\mathcal{H}_u$ for LLM fine-tuning to learn the personalized PEFT parameters. The training objectives for user $u$ under base, retrieval-augmented, and profile-augmented settings are:
\[
\left\{
\begin{aligned}
    \mathcal{L}_u^\mathrm{(B)}=&\mathrm{CE}[\Theta_u^\mathrm{(B)} (\phi_{t}(x_u)), y_u], \\
    \mathcal{L}_{u}^\mathrm{(R)} =& \mathrm{CE}[\Theta_u^\mathrm{(R)}(\phi_{r}(x_u, \mathcal{D}_u^{<t(x_u)})), y_u],\\
    \mathcal{L}_{u}^\mathrm{(P)} =& \mathrm{CE}[\Theta_u^\mathrm{(P)}(\phi_{p}(x_u, \mathcal{D}_u^{<t(x_u)}), s_u), y_u],
\end{aligned}
\right.
\]
where $\mathcal{D}_u^{<t(x_u)}=\mathcal{R}(\phi_t(x_u), \mathcal{H}_u^{<t(x_u)}, k)$, $\mathcal{H}_u^{<t(x_u)}$ is restricted to user $u$'s past behavior history that occurred before $x_u$.

User behavior history often does not align neatly with the query format. For example, in personalized tweet paraphrasing tasks, where the input is a text sequence $q_u$ and the output is the paraphrased tweet $r_u$, the history $\mathcal{H}_u$ only includes historical tweets. 
In scenarios where user history does not directly aligned with the specific task format, denoted as $\mathcal{H}_u=\{x_u\}$, we replace the user history output $y_u$ 
 in personal PEFT training objectives $\mathcal{L}_u^\mathrm{(B)}$, $\mathcal{L}_u^\mathrm{(R)}$, $\mathcal{L}_u^\mathrm{(P)}$ with right-shifted history $x_u'$ for unsupervised next token prediction.
 
 By optimizing personal PEFT parameters with the objectives mentioned above, \ourmethod{} comprehensively capture the user behavior patterns in PEFT parameters $\Delta\Theta_u^\mathrm{(B)}$, $\Delta\Theta_u^\mathrm{(R)}$, $\Delta\Theta_u^\mathrm{(P)}$, creating personalized LLMs owned by users. We envision the proposed \ourmethod{} as a versatile LLM personalization framework, where each user possesses their own PEFT parameters that contain personal behavior history and preferences. By plugging their personal PEFT parameters into the base LLMs, users can get their personalized LLMs, while achieving a better understanding and generalization of users' preferences from the parametric dimension.
 


\section{Experimental Settings}

\begin{table*}[t]
  \centering
    \caption{Main experiment results on the LaMP benchmark. R-1 and R-L denote ROUGE-1 and ROUGE-L, respectively. $k$ refers to the number of retrieved items, with $k=0$ indicating no retrieval. $\uparrow$ indicates that higher values are better, and $\downarrow$ implies lower values are preferable. For each task, the best score is in \textbf{bold} and the second best is \underline{underlined}.`$^*$' indicates significant improvement against counterparts without \ourmethod{}.}
  \begin{adjustbox}{max width=1\linewidth}
    \begin{tabular}{llccccccccccccc}
    \toprule[1.5pt]
    \multirow{2}{*}{\textbf{Task}} & \multirow{2}{*}{\textbf{Metric}} & \multicolumn{2}{c}{\textbf{Non-Personalized}} & \multicolumn{3}{c}{\textbf{RAG}} & \multicolumn{2}{c}{\textbf{PAG}} & \multicolumn{4}{c}{\textbf{RAG+\ourmethod{} (Ours)}} & \multicolumn{2}{c}{\textbf{PAG+\ourmethod{} (Ours)}} \\
    \cmidrule(r){3-4} \cmidrule(r){5-7} \cmidrule(r){8-9} \cmidrule(r){10-13} \cmidrule{14-15}
    & & k=0 & Random & k=1  & k=2 & k=4 & k=0  & k=1  & k=0 & k=1 & k=2 & k=4 & k=0 & k=1 \\
    \midrule[0.75pt]

    \multirow{2}{*}{\makecell[l]{\textsc{LaMP-1: Personalized}\\\textsc{Citation Identification}}} & Acc $\uparrow$ & .659 & .650 & .659 & .691 & .691 & .756 & .755 & .683$^*$ & .675$^*$ & .707$^*$ & .723$^*$ & \underline{.772}$^*$ & \textbf{.797}$^*$\\
    & F1 $\uparrow$ & .657 & .647 & .657 & .689 & .690 & .755 & .755 & .682$^*$ & .674$^*$ & .705$^*$ & .723$^*$ & \underline{.772}$^*$ & \textbf{.794}$^*$ \\
    \hline
    \multirow{2}{*}{\makecell[l]{\textsc{LaMP-2N: Personalized}\\\textsc{News Categorization}}} & Acc $\uparrow$& .787 & .785 & .820 & .832 & .832 & .817 & .817 & .810$^*$ & .823 & \underline{.834} & \textbf{.838}$^*$ & .827$^*$ & .831$^*$\\
    & F1 $\uparrow$& .538 & .527 & .598 & .632 & .647 & .623 & .621 & .589$^*$ & .615$^*$ & .635 & \textbf{.661}$^*$ & \underline{.648}$^*$ & .638$^*$ \\
    \hline
    \multirow{2}{*}{\makecell[l]{\textsc{LaMP-2M: Personalized}\\\textsc{Movie Tagging}}} & Acc $\uparrow$& .478 & .499 & .587 & .598 & .622 & .534 & .587 & .600$^*$ & .626$^*$ & .634$^*$ & \underline{.645}$^*$ & .636$^*$ & \textbf{.648}$^*$\\
    & F1 $\uparrow$& .425 & .441 & .512 & .514 & .542 & .476 & .506 & .493$^*$ & .531$^*$ & .535$^*$ & \textbf{.553}$^*$ & .536$^*$ & \underline{.540}$^*$\\
    \hline
    \multirow{2}{*}{\makecell[l]{\textsc{LaMP-3: Personalized}\\\textsc{Product Rating}}} & MAE $\downarrow$& .223 & .259 &.214 & .214 & .232 & .321 & .223 & \underline{.179}$^*$ &.196$^*$ & .214 & .223$^*$ & .205$^*$ & \textbf{.143}$^*$ \\
    & RMSE $\downarrow$& .491 & .590 & .535 & .463 & .535 & .582 & .473 & \underline{.443}$^*$ & .518$^*$ & .463 & .526$^*$ & .473$^*$ & \textbf{.378}$^*$\\
    \hline
    \multirow{2}{*}{\makecell[l]{\textsc{LaMP-4: Personalized}\\\textsc{News Headline Gen.}}} & R-1 $\uparrow$& .186 & .187 & .191 & .196 & \underline{.198} & .187 & .193 & .191$^*$ & .194$^*$ & .196 & \textbf{.199} & .189$^*$ & .194\\
    & R-L $\uparrow$& .167 & .168 & .172 & .176 & \underline{.178} & .168 & .173 & .171$^*$ & .175 & .177 & \textbf{.180}$^*$ & .170$^*$ & .175\\
    \hline
    \multirow{2}{*}{\makecell[l]{\textsc{LaMP-5: Personalized}\\\textsc{Scholarly Title Gen.}}} & R-1 $\uparrow$& .476 & .478 & .505 & .510 & .499 & .486 & .516 & .519$^*$ & .522$^*$ & .511 & \textbf{.526}$^*$ & .490$^*$ & \underline{.525}$^*$ \\
    & R-L $\uparrow$& .415 & .418 & .445 & .444 & .434 & .429 & .440 & .442$^*$ & .457$^*$ & .440 & \underline{.467}$^*$ & .428$^*$ & \textbf{.473}$^*$ \\
    \hline    \multirow{2}{*}{\makecell[l]{\textsc{LaMP-7: Personalized}\\\textsc{Tweet Paraphrasing}}} & R-1 $\uparrow$& .527 & .524 & .568 & .577 & .562 & .542 & .568 & .539$^*$ & \underline{.579}$^*$ & .575$^*$ & \textbf{.581}$^*$ & .542 & .577$^*$\\
    & R-L $\uparrow$& .474 & .474 & .521 & .527 & .514 & .501 & .518 & .483$^*$ & \textbf{.533}$^*$ & \underline{.531}$^*$ & .528$^*$ & .492 & \textbf{.533}$^*$\\
    \bottomrule[1.5pt]
    \end{tabular}%
    \end{adjustbox}
   
    \label{main_results}
\end{table*}

\begin{table}[t]
    \centering
    \caption{Dataset statistics: We report average sequence length in terms of number of tokens. \#Q is the number of queries, L$_{in}$ and L$_{out}$ are the average length of input and output sequence respectively, and \#History is the number of adopted items. To save space, task names can be found in Table~\ref{main_results}.}
    \begin{adjustbox}{max width=1\linewidth}
    \begin{tabular}{l r r r r r r r}
         \toprule[1.5pt]
        \multirow{2}{*}{\makecell[l]{\textbf{Task in}\\ \textbf{LaMP}}} & \multicolumn{3}{c}{\textbf{Base LLM Training}} & \multicolumn{4}{c}{\textbf{Personal PEFT Training}}\\
         \cmidrule(r){2-4} \cmidrule(r){5-8}
         &\#Q & L$_{in}$ & L$_{out}$ &  \#Q & \#History & L$_{in}$ & L$_{out}$ \\
         \midrule[0.75pt]
         \textbf{1}& 7,919 & 51.3 & 1.0 & 123 & 317.5 & 52.0 & 1.0\\
         \textbf{2M} & 3,181 &  92.1 & 1.4 & 3,302 &55.6 & 92.6 & 2.0\\
         \textbf{2N} & 3,662 & 68.2 & 1.3 & 6,033 & 219.9 & 63.5 & 1.1\\
         \textbf{3}& 22,388 & 128.7 & 1.0 & 112 & 959.8 & 211.9 & 1.0\\
         \textbf{4}& 7,275 & 33.9 & 9.2 & 6,275 & 270.1 & 25.2 & 11.1\\
         \textbf{5}& 16,075 & 162.1 & 9.7 & 107 & 442.9 & 171.6 & 10.3\\
         \textbf{7}& 14,826 & 29.7 & 18.3 & 109 & 121.2 & 29.4 & 18.0\\
         \bottomrule[1.5pt]
    \end{tabular}
    \end{adjustbox}
    \label{tab:benchmark_stat}
\end{table}

\paragraph{Datasets}
We use data from the Large Language Model Personalization (LaMP) benchmark \cite{salemi2023lamp}, which includes seven public language model personalization tasks: four classification tasks and three generation tasks.\footnote{We exclude LaMP-6 as it involves private data that we cannot access.} To promote LLM ownership, we emphasize the need for users to contribute extensive historical data for personalizing their model. Therefore, we focus on the most active users, selecting 100 users with the longest history logs from the time-based dataset version as the test set, while using all other users for base LLM training. 
Dataset statistics are presented in Table \ref{tab:benchmark_stat}.

\paragraph{Baselines}
We compare our proposed \ourmethod{} with the non-personalized baseline and the retrieval-augmented (RAG) and profile-augmented (PAG) LLM personalization methods.
For all baselines and \ourmethod{}, we choose one of the most widely adopted open-source LLM \texttt{Llama-2-7B} \cite{touvron2023llama} as our base LLM and take BM25 \cite{trotman2014improvements} for all retrieval operations to ensure efficient and fair comparison.\footnote{Baselines and hyperparameter details are presented in Appendix \ref{sec:baseline} and \ref{app:hyperparam} to facilitate further research.}

\paragraph{Evaluation Metrics}
Following LaMP \cite{salemi2023lamp}, we use accuracy and F1-score for classification tasks (LaMP-1, LaMP-2N, and LaMP-2M), MAE and RMSE for LaMP-3, and adopt ROUGE-1 and ROUGE-L \cite{lin-2004-rouge} for text generation tasks (LaMP-4, LaMP-5, LaMP-7). Note that all metrics are the higher the better, except for RMSE and MAE used for the LaMP-3.

\section{Results}
Table \ref{main_results} shows the performance on the test set for all seven public tasks in the LaMP benchmark, we have observations as follows.
\paragraph{\ourmethod{} brings universal improvement.} 
Models equipped with \ourmethod{} outperform all baseline personalization methods across all seven tasks. Notably, in personalized classification tasks, \ourmethod{} achieves an average relative improvement of 17.38\% in MAE and 8.89\% in RMSE for personalized product rating prediction. Additionally, it shows an 11.87\% improvement in accuracy and 7.56\% in F1-score for personalized movie tagging. For personalized text generation tasks, \ourmethod{} enhances ROUGE-1 and ROUGE-L scores by 3.42\% and 3.87\%, respectively, in personalized scholarly title generation.

\paragraph{Integrating non-parametric and parametric knowledge performs the best.} Combining \ourmethod{}'s parametric knowledge stored in PEFT parameters and the non-parametric in retrieved items and user profiles, results in notable performance gains. For instance, averaging across all seven tasks, combining retrieval in \ourmethod{} will bring 1.93\% and 2.48\% relative improvement compared with the non-retrieval and non-\ourmethod{} yet retrieval version model, respectively. Moreover, integrating \ourmethod{} with user profiles would also bring 4.56\% and 7.18\% performance gain against non-profile and non-\ourmethod{} versions, respectively. Overall, combining non-parametric retrieval and profile knowledge with parametric PEFT knowledge in \ourmethod{} delivers the best performance.


\paragraph{Performance \emph{w.r.t.} difference between task and history format.} 
In tasks like personalized citation identification, there is a notable discrepancy between the user history format and the task itself. Here, the user history comprises the user's publication history, while the task involves binary classification to identify the correct citation paper. This disparity is also seen in the personalized tweet paraphrasing task. In these cases, \ourmethod{} significantly enhances performance. Specifically, for personalized citation identification, \ourmethod{} increases accuracy by 3.48\% and F1-score by 3.52\%, thanks to personalized context knowledge provided through personal PEFT.
    

\paragraph{The more retrieved items, the better performance.} Our experimental results generally indicate that an increase in the number of retrieved items correlates with improved performance.
However, we also observe that some data points don't fit this trend, and we hypothesize that this inconsistency may arise from the retrieved items introducing noise and irrelevant behavior patterns, potentially complicating the model's process of understanding user preferences.


\begin{table}[t]
    \centering
    \caption{Performance under user behavior shift, where we remove the user behavior history highly similar to the query at hand. $k$ denotes the number of retrieved history items, and $k=0$ means non-retrieval. Armed with irrelevant user history, the retrieval-only method falls short and performs close to the non-personalized baseline, while \ourmethod{} shows stronger generalizability in the user behavior shift scenario.}
    \resizebox{1\linewidth}{!}{
    \begin{tabular}{l l c c c c c c c c}
        \toprule[1.5pt] 
         \makecell[l]{\textbf{LaMP}\\ \textbf{Task}}& \makecell[l]{\textbf{History}\\\textbf{Type}} &  \multicolumn{2}{c}{\makecell[c]{\textbf{Non-}\\\textbf{Personalized}}} & \multicolumn{2}{c}{\makecell[c]{\textbf{Retrieval}\\k=1}} & \multicolumn{2}{c}{\makecell[c]{\textbf{\ourmethod{}}\\k=0}} & \multicolumn{2}{c}{\makecell[c]{\textbf{\ourmethod{}}\\k=1}}  \\
        \cmidrule(r){1-10} 
        \multirow{3}{*}{1} &  & Acc & F1 & Acc & F1 & Acc & F1 & Acc & F1 \\
        \cmidrule(r){3-4} \cmidrule(r){5-6} \cmidrule(r){7-8} \cmidrule(r){9-10}
        & full & \multirow{2}{*}{.659}& \multirow{2}{*}{.657}&.659 & .657 & .683 & .682 & .675 & .674\\
        & irrelevant & & & .626 & .626 & .683 & .683 & .699 & .697 \\
        \cmidrule(r){1-10}
       \multirow{3}{*}{3} &  & MAE & RMSE & MAE & RMSE & MAE & RMSE  & MAE & RMSE\\
        \cmidrule(r){3-4} \cmidrule(r){5-6} \cmidrule(r){7-8}  \cmidrule(r){9-10}
        & full & \multirow{2}{*}{.223}& \multirow{2}{*}{.491} & .214 & .535 & .179 & .443 & .196 & .518 \\
        & irrelevant & & & .268 & .583 & .196 & .463 & .241 & .559 \\
        \cmidrule(r){1-10}
        \multirow{3}{*}{5} &  & R-1 & R-L & R-1 & R-L & R-1 & R-L & R-1 & R-L\\
        \cmidrule(r){3-4} \cmidrule(r){5-6} \cmidrule(r){7-8}  \cmidrule(r){9-10}
        & full & \multirow{2}{*}{.476}& \multirow{2}{*}{.415}& .505 & .445& .519 & .442 & .522  & .457\\
        & irrelevant & & & .475 & .417 & .493 & .437 & .490 & .417 \\
        \cmidrule(r){1-10}
       \multirow{3}{*}{7} &  & R-1 & R-L & R-1 & R-L & R-1 & R-L & R-1 & R-L\\
        \cmidrule(r){3-4} \cmidrule(r){5-6} \cmidrule(r){7-8}  \cmidrule(r){9-10}
        & full &\multirow{2}{*}{.527}  &\multirow{2}{*}{.474} & .571 & .521 & .539 & .483 & .579 & .533\\
        & irrelevant & & & .543 & .495 & .528 & .482 & .563 & .523 \\
       
       \bottomrule[1.5pt] 
    \end{tabular}
    }
    \label{tab:relevant_history}
\end{table}

\section{Analysis}
\paragraph{Performance under User Behavior Shift}

Recent studies have shown that retrieval-augmented generation methods tend to underperform when the retrieved corpus does not contain highly relevant documents \cite{shi2023large,gupta2024rag}. 
This problem is common in personalization contexts where the user's behavior history does not closely match their current queries. To simulate this scenario, we use \texttt{DeBERTa-v3} \cite{he2022debertav3} to extract features from the user's historical behaviors and current query, computing cosine similarity to assess relevance. We then rank the historical behaviors and select the top 100 items with the lowest relevance scores as irrelevant user history.

Table \ref{tab:relevant_history} shows that limiting user history to less relevant items significantly reduces the performance of retrieval-based methods, often aligning with non-personalized approaches. In contrast, \ourmethod{} demonstrates stronger robustness and generalization to less relevant history, even outperforming models trained with all user history items. Additionally, the combination of parametric and non-parametric knowledge (\ourmethod{}, $k$=1) enhances robustness in personalized text generation tasks, while models using only parametric knowledge (\ourmethod{}, $k$=0) perform better in personalized text classification tasks.

\begin{figure}[t]
    \centering
    \includegraphics[width=1\linewidth]{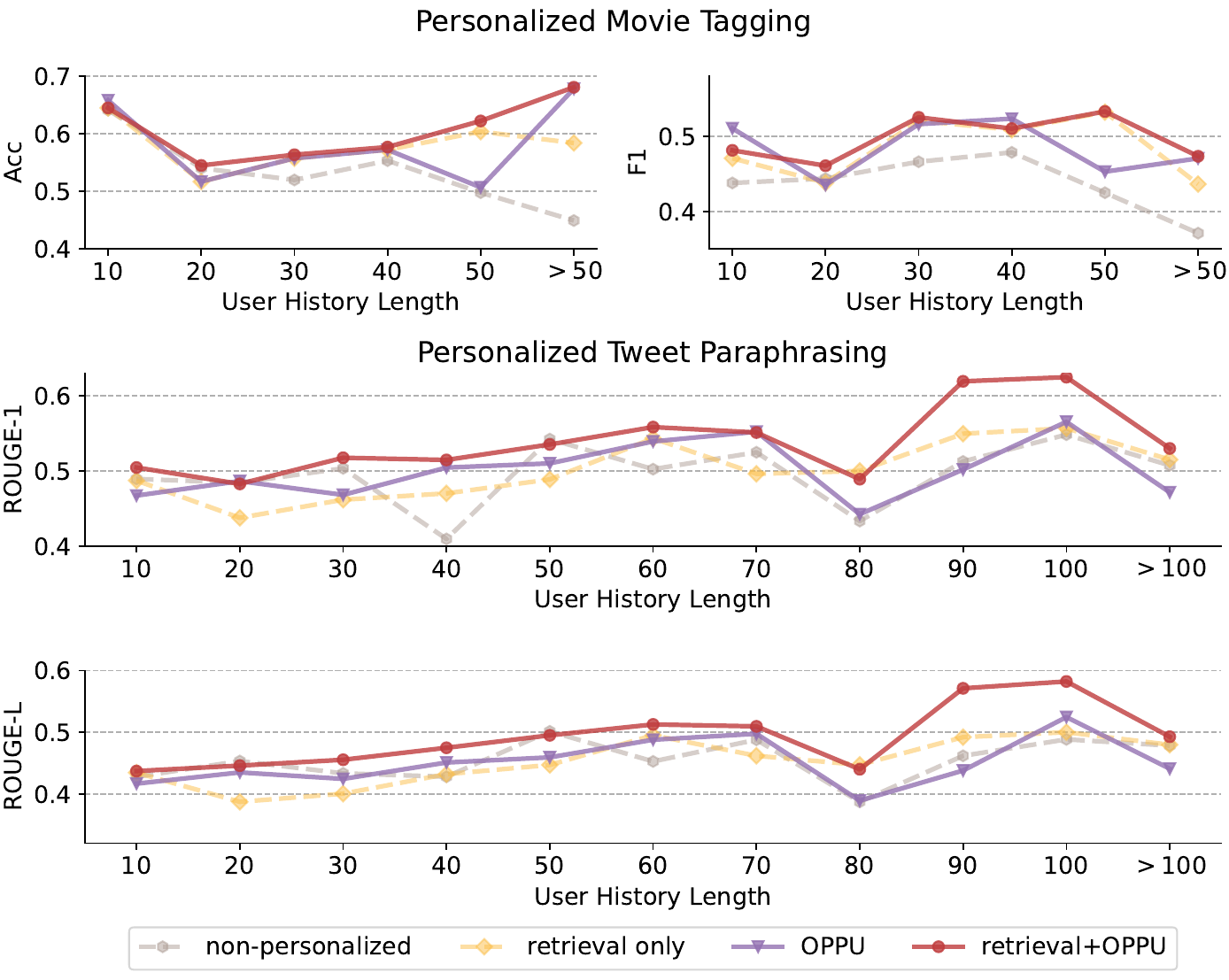}
    \caption{Model performance on personalized movie tagging and personalized tweet paraphrasing for users with different numbers of behavior history.}
    \label{fig:user_history_num}
\end{figure}

\paragraph{Modeling Users with Different Active Levels}

In our main experiment, we focus on highly active users. However, many users exhibit lower activity levels, resulting in shorter behavior histories. To examine the impact of user activity levels on model performance, we randomly selected 20 users from each activity range. Figure \ref{fig:user_history_num} shows that LLMs equipped with \ourmethod{} consistently outperform baseline methods across various activity levels. Key observations include: \textit{1)} The longer the user history, the more pronounced the superiority of retrieval + \ourmethod{} over baselines.
\textit{2)} Including non-parametric user knowledge via retrieval improves performance compared to methods without retrieval.
\textit{3)} Integrating parametric knowledge in \ourmethod{} with non-parametric knowledge from retrieval yields the strongest performance across different user activity levels.

\begin{figure}[t]
    \centering
    \includegraphics[width=0.95\linewidth]{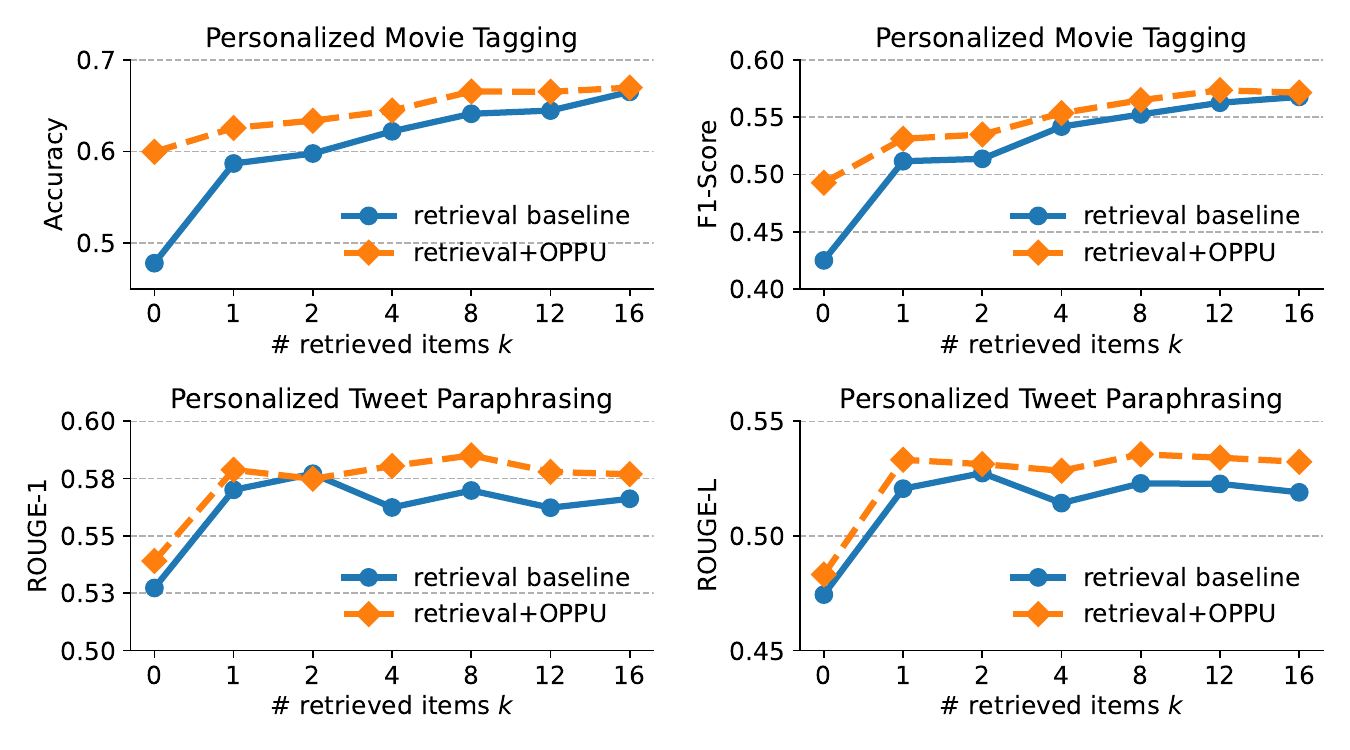}
    \caption{Performance of \ourmethod{} and retrieved-only baseline when the number of retrieved items $k$ increases.}
    \label{fig:retrieved_items}
\end{figure}

\paragraph{Performance \emph{w.r.t.} Retrieved History Items \textit{k}}

In this study, we alter the number of retrieved items of both retrieval-only baseline and retrieval+\ourmethod{} to gain a better understanding of the integration of non-parametric and parametric user knowledge. 
Figure \ref{fig:retrieved_items} illustrates that as we increase the number of retrieved historical behavior items, both the retrieval-only baselines and the retrieval+\ourmethod{} approaches show improved performance. Interestingly, we observe that as the number of retrieved items $k$ becomes larger, the performance difference between the retrieval-only and retrieval+\ourmethod{} narrows. This trend could be attributed to the longer logs of user behavior history in non-parametric prompts, which reduce the gap between the comprehensive user behavior history encapsulated in personalized PEFT parameters and the non-parametric user knowledge included in the prompts.

\begin{table}[t]
    \centering
    \caption{Performance of \ourmethod{} with different ablated versions of user history configurations. $k$ refers to the number of retrieved items, and $k=0$ denotes non-retrieval.
    The best score is in bold and the second best is \underline{underlined}.}
    \resizebox{1\linewidth}{!}{
    \begin{tabular}{l c c c c c c c c}
        \toprule[1.5pt] 
         \makecell[l]{\textbf{Task in}\\\textbf{LaMP}} & \multicolumn{2}{c}{\textbf{History}} & \multicolumn{2}{c}{\makecell[c]{\textbf{Retrieval}\\k=1}} & \multicolumn{2}{c}{\makecell[c]{\textbf{\ourmethod{}}\\k=0}} & \multicolumn{2}{c}{\makecell[c]{\textbf{\ourmethod{}}\\k=1}}  \\
        \cmidrule(r){1-3} \cmidrule(r){4-5} \cmidrule(r){6-7} \cmidrule(r){8-9}  
        \multirow{4}{*}{2M} & w/ desc. & w/ tag & Acc & F1 & Acc & F1 & Acc & F1 \\
        \cmidrule(r){2-9}
        & $\checkmark$ & & .530 & .488 & .486 & .437 & .624 & \underline{.539}\\
        & & $\checkmark$ & .567 & .514 & .499 &	.440 & \textbf{.634} & \textbf{.548} \\
        & $\checkmark$ & $\checkmark$ & .587 & .512 &	.600 &	.493 &	\underline{.626} &	.531\\
        \cmidrule(r){1-9}
        \multirow{4}{*}{5} & w/ abs. & w/ title & R-1 & R-L & R-1 & R-L & R-1 & R-L \\
        \cmidrule(r){2-9} 
        & $\checkmark$ & &  .493 & .422 & .497 &	.434 &	.495 &	\underline{.449}\\
        & & $\checkmark$ & .475 & .425 & .489 &	.430 & .492 &	.429 \\
        & $\checkmark$ & $\checkmark$ & .505& 	.445 &	\underline{.519} &	.442 &	\textbf{.522} & \textbf{.457}\\
       \bottomrule[1.5pt] 
    \end{tabular}
    }
    
    \label{tab:history_format}
\end{table}

\begin{figure*}[t]
    \centering
    \includegraphics[width=0.74\linewidth]{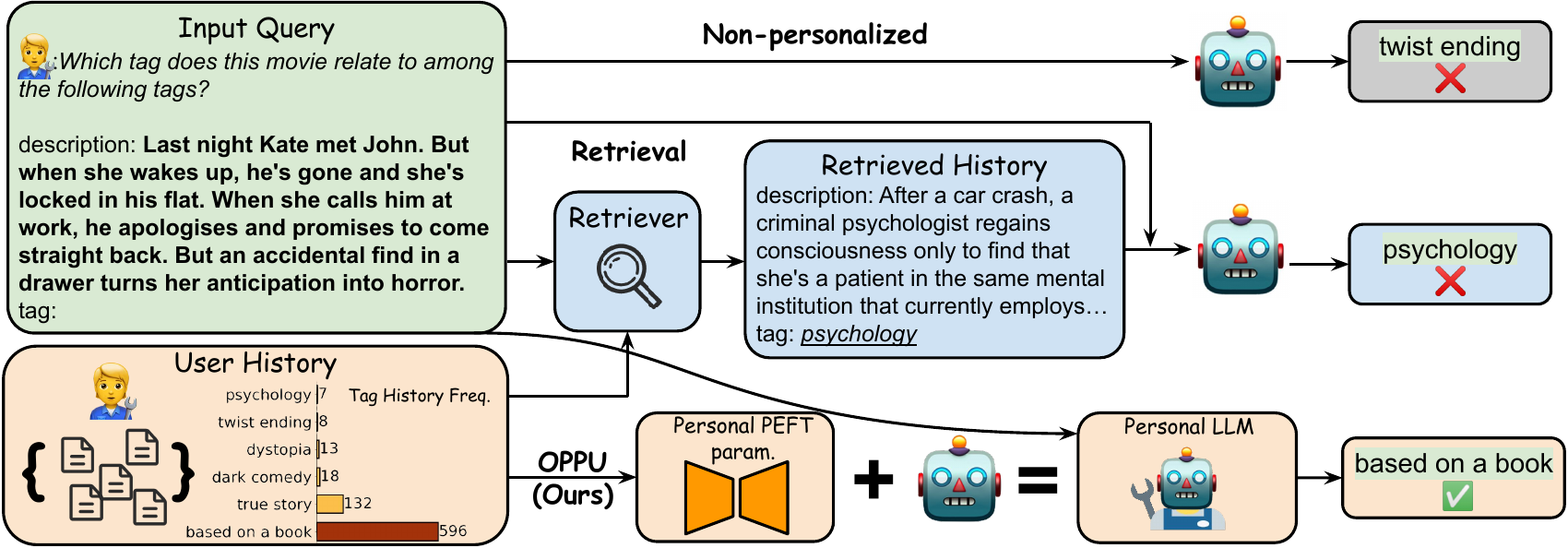}
    \caption{Case study in the personalized movie tagging task. It is shown that the retrieval-augmented personalization method can be easily distracted by less relevant user behavior history. In contrast, our \ourmethod{} demonstrates a more effective and comprehensive ability to capture the user's behavior patterns.}
    \label{fig:case_study}
\end{figure*}

\begin{figure}[t]
    \centering
    \includegraphics[width=1\linewidth]{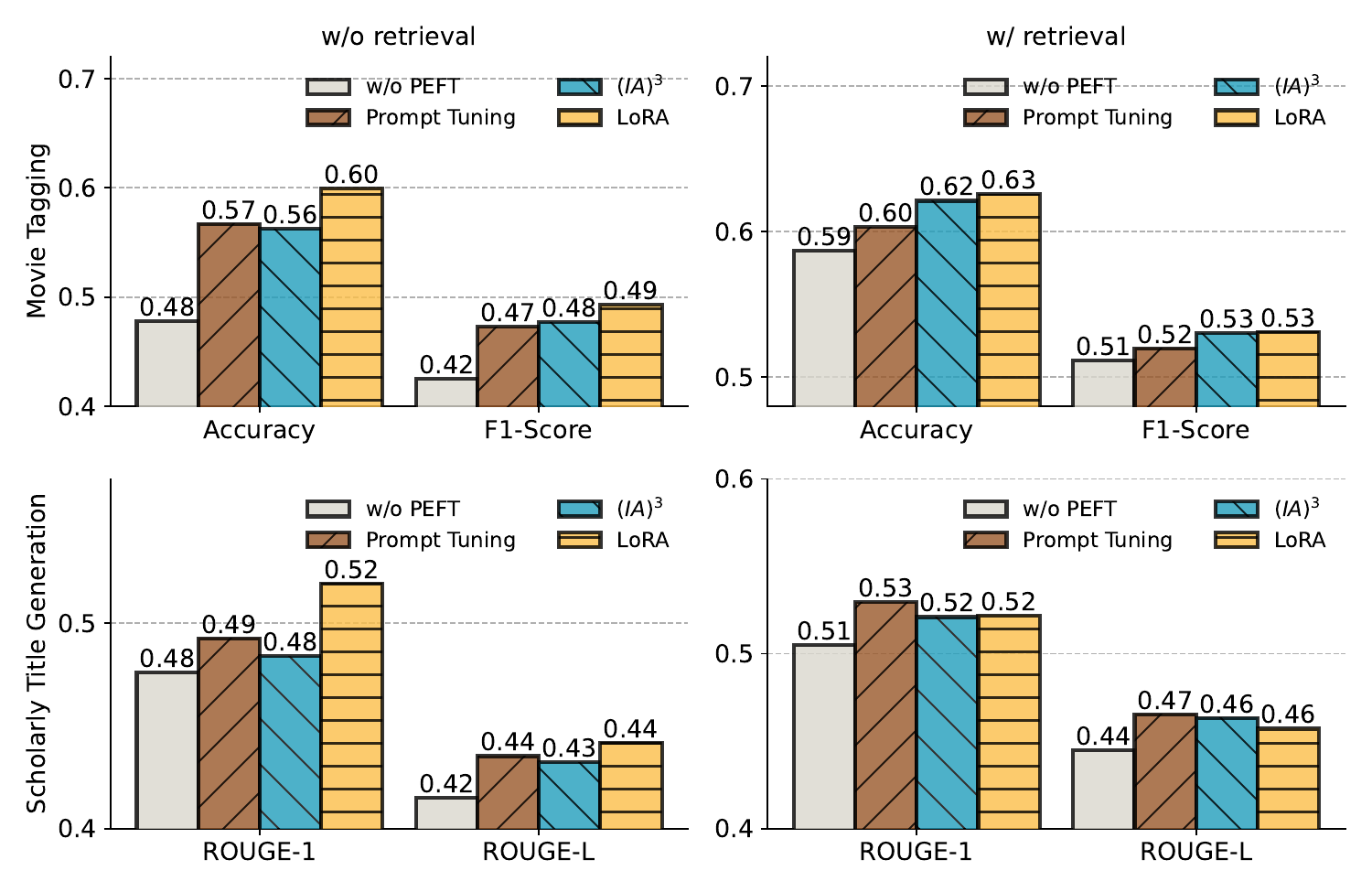}
    \caption{Performance of \ourmethod{} on personalized movie tagging and personalized scholarly title generation tasks when equipped with different PEFT methods. 
    We find that a larger proportion of trainable parameters generally results in better personalization performance.}
    \label{fig:peft_choice}
\end{figure}

\paragraph{Robustness against Task Formats}
Our main results demonstrate that \ourmethod{} significantly improves performance even when the user history corpus does not strictly follow the task format. We tested this robustness by ablating the history format in personalized movie tagging (LaMP-2M) and personalized scholarly title generation (LaMP-5) tasks, covering both text classification and generation categories.
In both tasks, each user history item consists of input and output aligned with the user query $x_u$ and output $y_u$. We ablated history behavior items from the input and output sides, comparing them with the retrieval baseline to test \ourmethod{}'s robustness against mismatched formats.

Shown in Table \ref{tab:history_format}, \ourmethod{} achieves performance close to that with full history in the text generation task, even with incomplete user behavior history. In news categorization, LLMs struggle with only parametric knowledge, but integrating retrieval augmentation, \ourmethod{} shows robust performance, outperforming models tuned on complete user history data. Overall, results reveal that combining non-parametric and parametric knowledge makes \ourmethod{} robust to different user history formats.

\paragraph{On PEFT Method Choices}
We propose \ourmethod{} as a versatile PEFT-based LLM personalization framework compatible with various PEFT methods. This study evaluates \ourmethod{}'s performance across different PEFT approaches, including LoRA, prompt tuning, and $\mathrm{(IA)^3}$, which plug in external learnable parameters in the embedding space and scale the attention factor, respectively.
As shown in Figure \ref{fig:peft_choice}, \ourmethod{} enhances performance with all three PEFT types, demonstrating its effectiveness and versatility. Notably, LoRA typically delivers the highest performance, followed by $\mathrm{(IA)^3}$, and then prompt tuning. This hierarchy aligns with the proportion of trainable parameters in each method: LoRA at 0.01\%, $\mathrm{(IA)^3}$ at 0.06\%, and prompt tuning at 0.001\%. These results suggest that a greater number of trainable parameters in a personalized PEFT method generally leads to improved personalization performance.

\paragraph{Case Study}
To illustrate the effectiveness of \ourmethod{}, we conduct a case study on personalized movie tagging task for an individual user. 
Figure \ref{fig:case_study} shows that the non-personalized method, relying solely on query input, ignores user behavior history and yields incorrect answers. The retrieval-based method, though incorporating user history, fails to retrieve closely matched behaviors to the query, also resulting in errors. We argue that retrieval augmentation with a few user history examples cannot fully capture user preferences. In contrast, \ourmethod{} uses a personalized PEFT module to effectively understand the user's behavior patterns across the entire user history. In this case, \ourmethod{} successfully recognizes the user's frequent tagging of ``based on a book" and provides the correct response.


\begin{figure*}[t]
    \centering
    \includegraphics[width=0.91\linewidth]{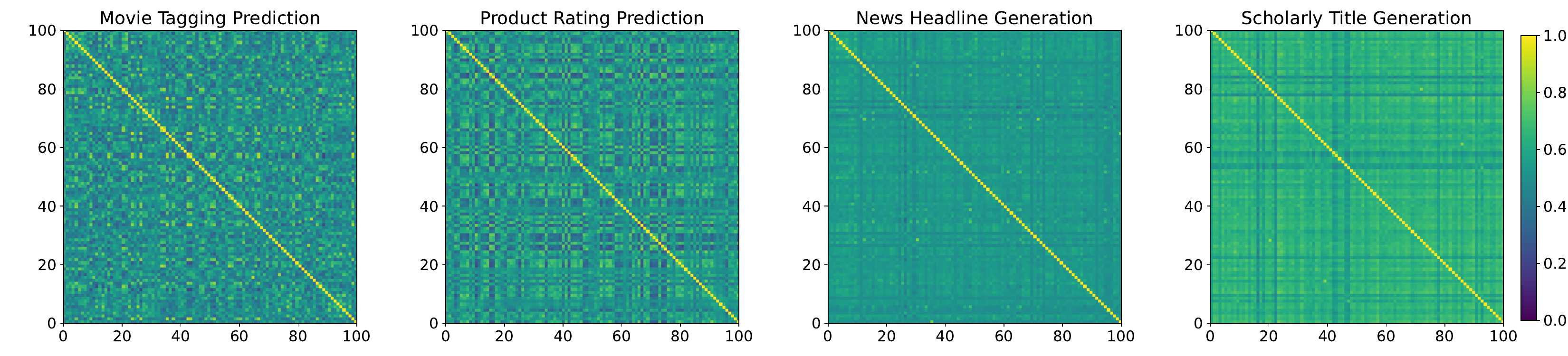}
    \caption{Similarities between personal PEFT parameters under personalized text classification and generation. }
    \label{fig:lora_similarity}
\end{figure*}

\paragraph{Similarities Between Personalized PEFTs}

To understand how user behavior patterns are reflected in their private PEFT parameters, we analyze the cosine similarities between these parameters across different users, as shown in Figure \ref{fig:lora_similarity}. We select two representative tasks from text classification and generation categories and compute the cosine similarities for 100 users' PEFT parameters in the test set. The private PEFT similarities generally range from 0.4 to 0.7, with the highest average similarities observed in the scholarly title generation task, likely due to its task-specific nature. Relative differences among users offer additional insights: in personalized text classification tasks, similarities vary more, indicating that some users have higher similarities than others. Conversely, in personalized text generation tasks, the similarities are relatively uniform, suggesting that personal preferences in these tasks are harder to categorize. 

        
    

\section{Related Work}


     

\subsection{Personalization of LLMs}
The thrust of existing LLM personalization research is centered on designing prompts that incorporate historical user-generated content and behavior. These approaches help LLMs understand users' preferences, tailoring responses to individual needs \cite{tan2023user, chen2023large}. The endeavors towards personalized LLMs mainly fall into three categories: {vanilla}, {retrieval-augmented}, and {profile-augmented personalized prompts}.

In the \textit{vanilla personalized prompt} category, researchers use in-context and few-shot learning to encode either complete or a sample of user behavior history as contextual examples \cite{liu2023chatgpt, wang2023learning}. For instance, \citet{dai2023uncovering} and \citet{kang2023llms} encode the user's personal rating history as few-shot demonstration examples. Moreover, some research works \cite{christakopoulou2023large, zhiyuli2023bookgpt} also discovered a long user history would bring better performance. To manage the growing user behavior data and LLMs’ limited context windows, the \textit{retrieval-augmented personalized prompt} approach has emerged \cite{salemi2023lamp, li2023teach}. For instance, Pearl \cite{mysore2023pearl} proposes a generation-calibrated retriever to select historic user-authored documents for prompt augmentation.
Beyond simple retrieval, some researchers summarize user preferences and behavior patterns into natural language profiles for input query augmentation, termed \textit{profile-augmented personalized prompts} \cite{liu2023once, sun2024persona}. \citet{richardson2023integrating} use the instruction-tuned LLMs to generate an abstract summary of user history data, augmenting retrieval-based personalization methods. There is also another line of work focusing on personalized alignment methods via parameter merging \cite{jang2023personalized} and personalized reward model \cite{cheng2023deserves}.



\subsection{Parameter-Efficient Fine-tuning (PEFT)}

With the exponential growth in LLM parameters, fine-tuning all parameters is expensive \cite{liu-etal-2022-p, xu2023parameter, gupta2024rag, jian-etal-2024-expedited}. To address this, parameter-efficient fine-tuning (PEFT) methods update only a small number of extra parameters while keeping pretrained weights frozen \cite{he2021towards, fu2023effectiveness, liu2024towards, dou2024avoiding, zhang-etal-2024-working-memory}. For example, adapter tuning \cite{houlsby2019parameter} injects learnable parameters into each feedforward layer, updating only these during fine-tuning. Inspired by discrete textual prompts \cite{sanh2022multitask, wang2022super}, prefix tuning \cite{li-liang-2021-prefix} and prompt tuning \cite{lester2021power} optimize prompts and prefixes for specific tasks. LoRA \cite{hu2021lora} adds low-rank matrices to approximate parameter updates, and $\rm{(IA)^3}$ \cite{liu2022few} scales activation in the attention mechanism. These methods achieve performance comparable to full fine-tuning by updating less than 1\% of the original parameters, are effective against catastrophic forgetting \cite{pfeiffer2021adapterfusion}, and are robust to out-of-distribution samples \cite{li-liang-2021-prefix}.

Previous works focused on prompt design, limited by model ownership and user behavior shifts. PEFT's small number of updated parameters and plug-and-play nature make it ideal for efficient LLM personalization and model ownership. \ourmethod{} introduces personalization at the parametric level via a personal PEFT module, pioneers storing user history within personal PEFT parameters, equipping each user with a unique, easily integrable PEFT module for model ownership.


\section{Conclusion}
We introduced \ourmethod{}, equipping each user with a personal PEFT module that facilitates model ownership and generalization under behavior shifts. By tuning these parameters with a user's history, \ourmethod{} captured personalized behavioral patterns. It integrated non-parametric user knowledge via retrieval and user profiles, showing superior performance across all seven LaMP benchmark tasks. Additional experiments demonstrated \ourmethod{}'s versatility, robustness, and effectiveness for users with varying activity levels. Our framework paved the way for new opportunities in PEFT-based LLM personalization, enhancing LLM modularity for effective and democratized personalization.

\section{Limitations}
We identify three key limitations in \ourmethod{}. 
Firstly, limited by the dataset, we mainly focus on one specific task per user rather than examining user behaviors across multiple tasks and domains. For example, in the movie tagging task, users are solely engaged in that specific activity, without the inclusion of behaviors from other areas. Despite this, the \ourmethod{} framework is inherently adaptable to any text sequence generation task and is capable of conducting diverse user instructions across different tasks and domains. The exploration of LLM personalization across a broader range of tasks and domains remains an area for future investigation.
Secondly, \ourmethod{} serves as a general framework that incorporates the entirety of a user's behavior history into their private PEFT module. However, user interests are dynamic and may display inconsistencies or conflicts over time. Future research directions include examining methodologies for selecting the most relevant or valuable items from a user's history and devising strategies to effectively manage any discrepancies or conflicts within this historical data. 

\section{Ethical Considerations}

\paragraph{Privacy}
Personalization in LLMs involves tailoring responses based on user-specific data, which may include sensitive or private information. The capacity of an LLM to adapt its outputs to individual users raises privacy concerns, as it might inadvertently reveal personal details. This underscores the importance of implementing robust privacy safeguards in LLM personalization, ensuring that personal data is handled respectfully and securely to prevent any unintended disclosures.

\paragraph{Data Bias}
Personalizing LLMs heavily relies on the personal data fed into the system. If this personal data is biased or unrepresentative, the model's outputs could potentially perpetuate these biases, leading to unfair or prejudiced responses. It is crucial to monitor and mitigate such biases in the personal data and the personalized model we obtain to ensure that personalized LLMs are fair and harmless in their responses.

\paragraph{Accessibility}
By advancing the field of LLM personalization, we aim to enrich user interactions with AI systems. However, the complexity and resource-intensive nature of LLMs might pose accessibility challenges. Smaller entities or individual researchers with limited computational power and budgetary constraints might find it difficult to engage with advanced personalized LLMs, potentially widening the gap in AI research and application. It is essential to develop strategies that make personalized LLM technologies more accessible to a broader range of users and researchers, ensuring equitable progress in this domain.

\section{Acknowledgements}

This work was supported by NSF IIS-2119531, IIS-2137396, IIS-2142827, IIS-2234058, CCF-1901059, and ONR N00014-22-1-2507.

\bibliography{custom}

\clearpage
\appendix

\begin{table}[t]
    \centering
    \caption{Hyperparameter settings of \ourmethod{} accross various tasks on LaMP benchmark. We find our hyperparameter settings robust across all 7 tasks.}
    \resizebox{1\linewidth}{!}{
    \begin{tabular}{l c c c c c}
        \toprule[1.5pt] 
        \textbf{Tasks} & \textbf{rank} & \textbf{\#epoch} & \textbf{lr} & \textbf{R2 reg.} & \textbf{batch size} \\
        \midrule[0.75pt] 
      \makecell[l]{\textsc{LaMP-1: Personalized}\\\textsc{Citation Identification}} & 8 &3& $1e^{-5}$ & $1e^{-2}$ & 16\\
      \hline
      \makecell[l]{\textsc{LaMP-2: Personalized}\\\textsc{News Categorization}} & 8 & 3 & $1e^{-5}$ & $1e^{-2}$ &  16\\
        \hline
      \makecell[l]{\textsc{LaMP-2: Personalized}\\\textsc{Movie Tagging}} & 8 & 3 & $1e^{-5}$ & $1e^{-2}$ & 4 \\
      \hline
      \makecell[l]{\textsc{LaMP-3: Personalized}\\\textsc{Product Rating}} & 8 & 3 & $1e^{-5}$ & $1e^{-2}$ & 3\\
      \hline
      \makecell[l]{\textsc{LaMP-4: Personalized}\\\textsc{News Headline Generation}} & 8  & 2 & $1e^{-5}$ & $1e^{-1}$ & 8\\
      \hline
      \makecell[l]{\textsc{LaMP-5: Personalized}\\\textsc{Scholarly Title Generation}} & 8 & 2 & $1e^{-5}$ & $1e^{-1}$ & 4\\
      \hline
      \makecell[l]{\textsc{LaMP-7: Personalized}\\\textsc{Tweet Paraphrasing}} & 8 & 2 & $1e^{-5}$ & $1e^{-1}$ & 8\\
       \bottomrule[1.5pt] 
    \end{tabular}
    }
    
    \label{tab:hyper}
\end{table}

\section{Hyperparameters}
\label{app:hyperparam}
The hyperparameters of \ourmethod{} are presented in Table \ref{tab:hyper} to facilitate further research. For LoRA, we add trainable low-rank matrice in the $W_q$ and $W_v$

\begin{figure}[t]
    \centering
    \includegraphics[width=1\linewidth]{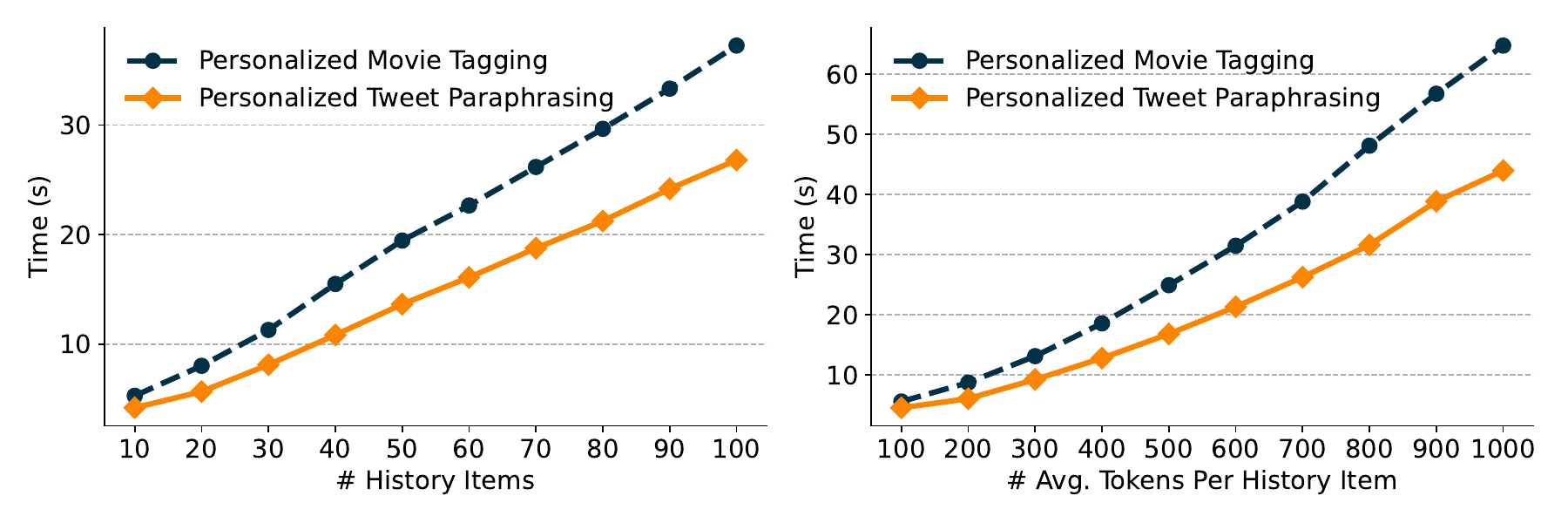}
    \caption{Efficiency analysis of \ourmethod{}, in which we alter the number of history items and average token per history item and record the training time. 
    }
    \label{fig:efficiency}
\end{figure}

\section{Efficiency Analysis}

Personalization is a technique that aims at universally benefiting everyone, where scalability and efficiency are crucial factors in large-scale deployment. In this experiment, we study the training efficiency of our proposed \ourmethod{}. We specifically examine two critical factors: the number of user history items and the average token numbers per history item across classification and generation tasks. Given that the training of each user's private PEFT can occur simultaneously or in a distributed manner, we choose not to consider the user count factor in this scenario, concentrating instead on the efficiency of training for an individual user. Initially, we set a consistent count of 100 whitespace-separated tokens for each history entry and vary the number of history items from 10 to 100. We then fix the history item count at 10 and adjust the token count from 10 to 100. The training time for each configuration, necessary for users to develop their personal PEFT modules. Presented in Figure \ref{fig:efficiency}, the results suggest that training time increases linearly with the number of user history items. Theoretically, training time grows quadratically with the increase in average tokens per history entry, yet our observations indicate a trend more akin to linear growth. It's noteworthy that the longer training durations for personalized movie tagging tasks, as opposed to personalized tweet paraphrasing, are attributed to different training epochs.

\section{Baseline Details}
\label{sec:baseline}
The baseline details are presented as follows:
\begin{itemize}[leftmargin=*]
    \item \textbf{Non-Personalized Baseline}: We present two approaches under the non-personalized setting: non-retrieval and random history. \textit{Non-retrieval method} refers to only feeding the user's query without revealing the user's behavior history to the LLMs. \textit{Random history} baseline means augmenting the user's query with random history behavior from all user history corpus.

    \item \textbf{Retrieval-Augmented Personalization (RAG)}: We follow the retrieval-augmented personalization method presented in LaMP \cite{salemi2023lamp}, where the user's query is augmented with top $k$ retrieved items from the corresponding user's history corpus. We take $k$=1, 2, 4 in this work.

    \item \textbf{Profile-Augmented Personalization (PAG)}: This method is taken from \citet{richardson2023integrating}, in which the user's input sequence would concatenate the user's profile summarizing the user's preference and behavior patterns. In our experiments, we generate user profiles using the \texttt{vicuna-7B} \cite{chiang2023vicuna} model. Moreover, the profile-augmented method could be combined with the retrieval augmentation. In this case, we take the number of retrieval items $k$=1 following the setting of \citet{richardson2023integrating}.
\end{itemize}

\section{Scientific Artifacts}
\ourmethod{} is built with the help of many existing scientific artifacts, including PyTorch \cite{paszke2019pytorch}, Numpy \cite{harris2020array}, huggingface transformers \cite{wolf2020transformers}, and bitsandbytes \cite{dettmers2022llmint8}. We will make the \ourmethod{} implementation publicly available to facilitate further research.

\section{Computation Resources Details}
All experiments are implemented on a server with 3 NVIDIA A6000 GPU and Intel(R) Xeon(R) Silver 4210R CPU @ 2.40GHz with 20 CPU cores. Training 100 personal PEFT sequentially took around 12 minutes to 12 hours depending on the size of the behavior history corpus and the sequence length per history item.

\section{PEFT Cosine Similarity Details}
Each user's private PEFT parameters contain multiple learnable tensors, we first flatten the tensors and calculate the cosine similarities between corresponding private PEFT parameters, then average cosine similarities for each pair of PEFT modules. A pseudo-code using PyTorch is as follows:
\begin{verbatim}
def cosine_similarity(PEFT_1, PEFT_2):
    similarity_sum = 0
    count = 0    
    for key in PEFT_1:
        if key in PEFT_2:
            v1 = PEFT_1[key].flatten()
            v2 = PEFT_2[key].flatten()                
       
            dot = torch.dot(v1, v2)
            norm_1 = torch.linalg.norm(v1)
            norm_2 = torch.linalg.norm(v2)    
            
            similarity = dot / (norm_1 * norm_2)
            similarity_sum += similarity
            count += 1        
    
    return similarity_sum / count
\end{verbatim}

\section{Task Details}
We present the task details as follows to help readers gain a better understanding of the task format.
\begin{itemize}[leftmargin=*]
    \item \textbf{Personalized Citation Identification} is a binary text classification task. Specifically, given user $u$ writes a paper $x$, the task aims to make the model determine which of the two candidate papers $u$ will cite in paper $x$ based on the user's history data, which contains the publications of user $u$.

    \item \textbf{Personalized News Categorization} is a 15-way text classification task to classify news articles written by a user $u$. Formally, given a news article $x$ written by user $u$, the language model is required to predict its category from the set of categories based on the user's history data, which contains the user's past article and corresponding category.

    \item \textbf{Personalized Movie Tagging} is a 15-way text classification task to make tag assignments aligned with the user's history tagging preference. Specifically, given a movie description $x$, the model needs to predict one of the tags for the movie $x$ based on the user's historical movie-tag pairs.

    \item \textbf{Personalized Product Rating} is a 5-way text classification task and can also be understood as a regression task. Given the user $u$'s historical review and rating pairs and the input review $x$, the model needs to predict the rating corresponding to $x$ selected from 1 to 5 in integer. 

    \item \textbf{Personalized News Headline Generation} is a text generation task to test the model's ability to capture the stylistic patterns in personal data. Given a query $x$ that requests to generate a news headline for an article, as well as the user profile that contains the author's historical article-title pairs, the model is required to generate a news headline specifically for the given user.

    \item \textbf{Personalized Scholarly Title Generation} is a text generation task to test personalized text generation tasks in different domains. In this task, we require language models to generate titles for an input article $x$, given a user profile of historical article-title pairs for an author.

    \item \textbf{Personalized Tweet Paraphrasing} is also a text generation task that tests the model's capabilities in capturing the stylistic patterns of authors. Given a user input text $x$ and the user profile of historical tweets, the model is required to paraphrase $x$ into $y$ that follows the given user's tweet pattern.
    
\end{itemize}

\section{Prompt for Personalization Tasks}
We present the prompt used in our experiments in this section, where the text in \texttt{\{BRACES\}} can be replaced with content specific to different users and queries.

\subsection{Personalized Citation Identification}

\texttt{\{USER PROFILE\}}

\noindent\texttt{\{RETRIEVED HISTORY\}}

\noindent Identify the most relevant reference for the listed publication by the researcher. Select the reference paper that is most closely related to the researcher\'s work. Please respond with only the number that corresponds to the reference.

\noindent paper title: \texttt{\{QUERY PAPER TITLE\}}

\noindent reference: [1] - \texttt{\{OPTION1\}} [2] - \texttt{\{OPTION2\}}

\noindent answer:

\subsection{Personalized News Categorization}
\texttt{\{USER PROFILE\}}

\noindent\texttt{\{RETRIEVED HISTORY\}}

\noindent Which category does this article relate to among the following categories? Just answer with the category name without further explanation. categories: [travel, education, parents, style \& beauty, entertainment, food \& drink, science \& technology, business, sports, healthy living, women, politics, crime, culture \& arts, religion]

\noindent article: \texttt{\{QUERY ARTICLE\}} category:

\subsection{Personalized Movie Tagging}
\texttt{\{USER PROFILE\}}

\noindent\texttt{\{RETRIEVED HISTORY\}} 

\noindent Which tag does this movie relate to among the following tags? Just answer with the tag name without further explanation. tags: [sci-fi, based on a book, comedy, action, twist ending, dystopia, dark comedy, classic, psychology, fantasy, romance, thought-provoking, social commentary, violence, true story]

\noindent description: \texttt{\{QUERY DESCRIPTION\}} tag:

\subsection{Personalized Product Rating}
\texttt{\{USER PROFILE\}}

\noindent\texttt{\{RETRIEVED HISTORY\}}

\noindent What is the score of the following review on a scale of 1 to 5? just answer with 1, 2, 3, 4, or 5 without further explanation. 

\noindent review: \texttt{\{QUERY REVIEW\}} score:

\subsection{Personalized News Headline Generation}
\texttt{\{USER PROFILE\}}

\noindent\texttt{\{RETRIEVED HISTORY\}}

\noindent Generate a headline for the following article.

\noindent article: \texttt{\{QUERY ARTICLE\}} headline:

\subsection{Personalized Scholarly Title Generation}
\texttt{\{USER PROFILE\}}

\noindent\texttt{\{RETRIEVED HISTORY\}}

\noindent Generate a title for the following abstract of a paper.

\noindent abstract: \texttt{\{QUERY ABSTRACT\}} title:

\subsection{Personalized Tweet Paraphrasing}
\texttt{\{USER PROFILE\}}

\noindent\texttt{\{RETRIEVED HISTORY\}}

\noindent Following the given pattern, paraphrase the following text into tweet without any explanation before or after it.

\noindent text: \texttt{\{QUERY TEXT\}} tweet:

\section{Prompt for User Profile Generation}
For user profile generation, we follow the prompt template in \citet{richardson2023integrating}. 

\subsection{Personalized Citation Identification}
Write a summary, in English, of the research interests and topics of a researcher who has published the following papers. Only generate the summary, no other text. User History: \texttt{\{USER HISTORY\}} Answer:

\subsection{Personalized News Categorization}
Look at the following past articles this journalist has written and determine the most popular category they write in. Answer in the following form: most popular category: <category>. User History: \texttt{\{USER HISTORY\}} Answer:

\subsection{Personalized Movie Tagging}
Look at the following past movies this user has watched and determine the most popular tag they labeled. Answer in the following form: most popular tag: <tag>. User History: \texttt{\{USER HISTORY\}} Answer:

\subsection{Personalized Product Rating}
Based on this user’s past reviews, what are the most common scores they give for positive and negative reviews? Answer in the following form: most common positive score: <most common positive score>, most common negative score: <most common negative score>. User History: {} Answer:Look at the following past movies this user has watched and determine the most popular tag they labeled. Answer in the following form: most popular tag: <tag>. User History: \texttt{\{USER HISTORY\}} Answer:

\subsection{Personalized News Headline Generation}
Given this author’s previous articles, try to describe a template for their headlines. I want to be able to accurately predict the headline gives one of their articles. Be specific about their style and wording, don’t tell me anything generic. User History: \texttt{\{USER HISTORY\}} Answer:

\subsection{Personalized Scholarly Title Generation}
Given this author’s previous publications, try to describe a template for their titles. I want to be able to accurately predict the title of one of the papers from the abstract. Only generate the template description, nothing else. User History: \texttt{\{USER HISTORY\}} Answer:

\subsection{Personalized Tweet Paraphrasing}
Given this person’s previous tweets, try to describe a template for their tweets. I want to take a generic sentence and rephrase it to sound like one of their tweets, with the same style/punctuation/capitalization/wording/tone/etc. as them. Only give me the template description, nothing else. User History: \texttt{\{USER HISTORY\}} Answer:

\end{document}